\documentclass[amsthm]{elsart}
\usepackage{yjsco}
\usepackage{natbib}

\usepackage{algorithm}
\usepackage{algpseudocode}
\usepackage{graphicx}
\usepackage{url}
\usepackage{xspace}
\usepackage{hyperref}
\usepackage{color}
\usepackage{underscore}
\usepackage{tipa}
\usepackage{cite}
\usepackage{stmaryrd}
\usepackage{amsmath}
\usepackage{amssymb}
\usepackage{booktabs}
\usepackage{listings}
\usepackage{paralist}
\usepackage{subfig}
\usepackage{hyperref}
\usepackage{breakurl}
\usepackage{tikz}
\usepackage{textcomp}
\usepackage{alltt}
\usepackage{amsmath}

\renewcommand{\arraystretch}{1.2}

\algrenewcommand\algorithmicrequire{\textbf{Input}}
\algrenewcommand\algorithmicensure{\textbf{Output}}

\usepackage{footnote}

\makeatletter

\def\holdocspecials{\do\ \do\$\do\&%
  \do\#\do\^\do\^^K\do\_\do\^^A\do\%}

\def\holtt{\trivlist \item[]\if@minipage\else\vskip\parskip\fi
\leftskip\@totalleftmargin\rightskip\z@
\parindent\z@\parfillskip\@flushglue\parskip\z@
\@tempswafalse \def\par{\if@tempswa\hbox{}\fi\@tempswatrue\@@par}
\obeylines \tt \let\do\@makeother \holdocspecials
 \frenchspacing\@vobeyspaces}

\makeatother

\newlength{\hsbw}
\newcommand\HOLSpacing{13pt}

\newenvironment{holnb}{\begin{flushleft}
 \begin{minipage}[b]{\hsbw}
 \vspace*{.06in}
 \begingroup\small\baselineskip\HOLSpacing\footnotesize
 \begin{holtt}}{\end{holtt}\endgroup
 \end{minipage}
 \end{flushleft}}

   \newcommand\hilbert{\varepsilon}
   
   \newcommand{\Thm}{\(\vdash\)}
   \newcommand{\Cond}{\(\rightarrow\)}
   \newcommand{\Eqv}{\(\equiv\)}
   \newcommand{\Iff}{\(\Longleftrightarrow\)\hspace{-1.5mm}}
   \newcommand{\Fa}{\(\forall\)}
   \newcommand{\Et}{\(\exists\)}
   \newcommand{\Eu}{\(\exists_{unique}\)}

   \newcommand{\Impl}{\(\Longrightarrow\)\hspace{-1.5mm}}
   \newcommand{\Func}{\(\to\)\hspace{-1.5mm}}

   \newcommand{\Lam}{\(\lambda\)}
   
   \newcommand{\Minus}{\(-\)}
   \newcommand{\Lminus}{\(-\)\hspace{-1.5mm}}
   \newcommand{\Prime}{\('\)}
   \newcommand{\Und}{\_}
   \newcommand{\Lt}{\(<\)}
   \newcommand{\Gt}{\(>\)}
   \newcommand{\Leq}{\(\leq\)}
   \newcommand{\Geq}{\(\geq\)}
   \newcommand{\Eq}{\(=\)}
   \newcommand{\Lrb}{\((\)}
   \newcommand{\Rrb}{\()\)}

   \newcommand{\Next}{\(\bigcirc\)}
   \newcommand{\Prev}{\(\ominus\)}
   \newcommand{\WPrev}{\(\widetilde{\bigcirc}\)}
   \newcommand{\Event}{\(\Diamond\)}
   \newcommand{\Once}{\(\underline{\Diamond}\)}  
\newcommand{\Hilbert}{\(\hilbert\)}

\newcommand{\Conj}{\(\wedge\)}
\newcommand{\Disj}{\(\vee\)}
\newcommand{\Neg}{\(\neg\)}
\newcommand{\Pnd}{\(\Diamond\)}

\newcommand{\Models}{\(\models\)}

\long\def\holthm#1{{\def\Turns{\Thm} \rechol#1\end\end\end}}

\long\def\rechol#1#2#3{\let\next=\rechol\def\postnext{#2#3}\ifx#1\end
\let\next=\relax\def\postnext{\relax}
\else\ifx#1!\Fa                                          
\else\ifx#1@\Hilbert                                     
\else\ifx#1\#\Pnd                                        
\else\ifx#1'\Prime                                       
\else\ifx#1~\Neg                                         
\else\ifx#1\~\Neg
\else\ifx#1_\Und                                         
\else\ifx#1(\ifx#2+\ifx#3)\Next\def\postnext{}\fi        
            \else\ifx#2-\Prev\def\postnext{}             
            \else\ifx#2~\ifx#3)\WPrev\def\postnext{}\fi            
             \else\Lrb\fi\fi\fi                          
\else\ifx#1)\Rrb%
\else\ifx#1\/\Disj                                       
\else\ifx#1\.\Lam                                        
\else\ifx#1>\ifx#2=\Geq\def\postnext{#3}\else\Gt\fi      
\else\ifx#1?\ifx#2!\Eu\def\postnext{#3}\else\Et\fi       
\else\ifx#1-\ifx#2>\Func\def\postnext{#3}               
            \else\ifx#2-\Lminus\def\postnext{#3}
            \else\Minus\fi\fi                               
\else\ifx#1|\ifx#2-\Turns\def\postnext{#3}               
            \else\ifx#2=\Models\def\postnext{#3}
                 \else\Bar\fi\fi
\else\ifx#1<\ifx#2=\ifx#3>\Iff\def\postnext{}       
                   \else\Leq\def\postnext{#3}\fi    
            \else\ifx#2+\Event\def\postnext{}       
            \else\ifx#2-\Once\def\postnext{}       
            \else\Lt\fi\fi\fi                       
\else\ifx#1=\ifx#2=\ifx#3>\Impl\def\postnext{}            
                   \else\Eqv\def\postnext{#3}\fi         
            \else\ifx#2>\Cond\def\postnext{#3}
                 \else\Eq\fi\fi
\else\ifx#1/\ifx#2\^^M\Conj\par\def\postnext{#3}         
            \else\ifx#2\ \Conj\ \def\postnext{#3}\else#1\fi\fi  
\else#1\fi\fi\fi\fi\fi\fi\fi\fi\fi\fi\fi\fi\fi\fi\fi\fi\fi\fi\fi
\expandafter\next\postnext}

\def\systemname#1{\textsf{#1}\xspace}
\newcommand{\coloneqq}[0]{\mathrel{\mathop:}=}
\newcommand{\HOLLight}{\systemname{HOL Light}}
\newcommand{\HOL}{\systemname{HOL}}

\newcommand{\Isabelle}{\systemname{Isabelle}}
\newcommand{\Sledgehammer}{\systemname{Sledgehammer}}
\newcommand{\MaSh}{\systemname{MaSh}}

\newcommand{\Mizar}{\systemname{Mizar}}

\newcommand{\MizAR}{\systemname{Miz$\mathbb{AR}$}}

\newcommand{\Flyspeck}{\systemname{Flyspeck}}

\newcommand{\MoMM}{\systemname{MoMM}}

\newcommand{\OCaml}{\systemname{OCaml}}

\newcommand{\HH}{\systemname{HOL\hspace{-.5mm}\textsc{\raisebox{.5mm}{\scriptsize y}}Hammer}}
\newcommand{\epcllemma}{\systemname{epcllemma}}
\newcommand{\AGIntRater}{\systemname{AGIntRater}}
\newcommand{\Agint}{\systemname{AGInt}}

\newcommand{\MESON}{\texttt{MESON}\xspace}

\begin{document}

\begin{frontmatter}
\title{Learning-assisted Theorem Proving with Millions of Lemmas}
\thanks{This research was partly supported by FWF grant P26201}

\author{Cezary Kaliszyk}
\address{University of Innsbruck, Austria}
\ead{cezary.kaliszyk@uibk.ac.at}
\ead[url]{\url{http://cl-informatik.uibk.ac.at/~cek/}}

\author{Josef Urban}
\address{Radboud University, Nijmegen}
\ead{josef.urban@gmail.com}
\ead[url]{\url{http://cs.ru.nl/~urban/}}

\begin{abstract}
  Large formal mathematical libraries consist of millions of atomic
  inference steps that give rise to a corresponding number of proved
  statements (lemmas). Analogously to the informal mathematical
  practice, only a tiny fraction of such statements is named and
  re-used in later proofs by formal mathematicians. In this work, we
  suggest and implement criteria defining the estimated usefulness of
  the \HOLLight lemmas for proving further theorems. We use these
  criteria to mine the large inference graph of the lemmas in the \HOLLight
  and \Flyspeck libraries, adding up to millions of the best
  lemmas to the pool of statements that can be re-used in later
  proofs. We show that in combination with learning-based relevance
  filtering, such methods significantly strengthen automated theorem
  proving of new conjectures over large formal mathematical libraries
  such as \Flyspeck. 

\end{abstract}


\end{frontmatter}

\section{Introduction: Automated Reasoning over Large Mathematical Libraries}
\label{Introduction}

In the last decade, large formal mathematical corpora such as the
Mizar Mathematical Library~\citep{mizar-in-a-nutshell} (MML), \Isabelle/\HOL~\citep{WenzelPN08} and \HOLLight~\citep{Harrison96}/\Flyspeck~\citep{Hales05}
have been translated to formats that allow easy experiments with
external automated theorem provers (ATPs) 
and AI systems~\citep{Urb04-MPTP0,MengP08,holyhammer}. 

The problem that has immediately emerged is to efficiently
automatically reason over such large formal mathematical knowledge
bases, providing as much support for authoring computer-understandable
mathematics as possible.  Reasoning with and over such large ITP
(interactive theorem proving) libraries is however not just a new
problem, but also a new opportunity, because the libraries already
contain a lot of advanced knowledge in the form of concepts, theorems,
proofs, and whole theory developments. Such large pre-existing
knowledge allows mathematicians to state more advanced conjectures,
and experiment on them with the power of existing symbolic reasoning
methods. The large amount of mathematical and problem-solving
knowledge contained in the libraries can be also subjected to all
kinds of knowledge-extraction methods, which can later complement more
exhaustive theorem-proving methods by providing domain-specific
guidance. Developing the strongest possible symbolic reasoning methods
that combine such knowledge extraction and re-use with correct
deductive search is an exciting new area of Artificial Intelligence
and Symbolic Computation.

Several symbolic AI/ATP methods for reasoning in the context of a
large number of related theorems and proofs have been suggested and
tried already, including: (i) methods (often external to the core
ATP algorithms) that select relevant premises (facts)
from the thousands of theorems available in such corpora~\citep{HoderV11,KuhlweinLTUH12}, (ii) methods
for internal guidance of ATP systems when reasoning in the large-theory
setting~\citep{UrbanVS11}, (iii) methods that automatically evolve more and more
efficient ATP strategies for the clusters of related problems from
such corpora~\citep{blistr}, and (iv) methods that learn which of such specialized strategies to use
for a new problem~\citep{KuhlweinSU13}.

In this work, we start to complement the first set of methods --
ATP-external premise selection -- with \textit{lemma mining} from the large corpora. The main
idea of this approach is to enrich the pool of human-defined main
(top-level) theorems in the large libraries with the most
useful/interesting lemmas extracted from the proofs in these
libraries. Such lemmas are then eligible together with (or instead of)
the main library theorems as the premises that are given to the ATPs
to attack new conjectures formulated over the large libraries. 

This high-level
idea is straightforward, but there are a number of possible
approaches involving a number of issues to be solved, starting 
with a reasonable definition of a \textit{useful/interesting lemma},
and with making such definitions efficient over corpora that contain
millions to billions of candidate lemmas. These issues are discussed
in Sections~\ref{Data} and \ref{Good}, after motivating and explaining the overall approach 
for using lemmas in large theories in Section~\ref{Approach} and giving an overview of the 
recent related work in Section~\ref{Related}.

As in any AI discipline dealing with large amount of data, research in
the large-theory field is driven by rigorous experimental evaluations
of the proposed methods over the existing corpora. 
For the first experiments with lemma
mining we use the \HOLLight system, together with its core library and the \Flyspeck library.
The various evaluation scenarios are defined and discussed in Section~\ref{Scenarios},
and the implemented methods are evaluated
in Section~\ref{Experiments}.
Section~\ref{Future} discusses the
various future directions and concludes.\footnote{This paper is an extended version of~\citep{KaliszykU13a}.}

\section{Using Lemmas for Theorem Proving in  Large Theories}
\label{Approach}
The main task in the Automated Reasoning in Large Theories (ARLT) domain is to prove new
conjectures with the knowledge of a large body of previously proved
theorems and their proofs. This setting reasonably corresponds to how
large ITP libraries are constructed, and hopefully also emulates how human mathematicians work more
faithfully than the classical scenario
of a single hard problem consisting of isolated axioms and a
conjecture~\citep{UrbanV13}. The pool of previously proved theorems
ranges from thousands in large-theory ATP benchmarks such as MPTP2078~\citep{abs-1108-3446},
to tens of thousands when working with the whole ITP
libraries.\footnote{23323 theorems are in the \HOLLight/\Flyspeck library (SVN revision 3437), about 20000 are in the \Isabelle/\HOL library, and
  about 50000 theorems are in the \Mizar library.}

The strongest existing ARLT systems combine variously parametrized
premise-selection techniques (often based on machine learning from previous proofs) with  
ATP systems and their strategies that are called with varied
numbers of the most promising premises. These techniques can go quite
far already: when using 14-fold parallelization and 30s wall-clock
time, the \HH system~\citep{holyhammer,KaliszykU13} can today prove
47\% of the 14185\footnote{These experiments were done on a earlier version of 
\Flyspeck (SVN revision 2887) than is used here (SVN revision 3437), where the number of 
theorems is 23323.} \Flyspeck theorems~\citep{EasyChair:74}. This is
measured in a scenario\footnote{A similar scenario has been 
  introduced in 2013 also for the LTB (Large-Theory Batch) division of the CASC competition.} in which the \Flyspeck
theorems are ordered \textit{chronologically} using the loading
sequence of the \Flyspeck library, and presented in this order to
\HH as conjectures. After each theorem is attempted, its human-designed \HOLLight
proof is fed to the \HH's learning components, together
with the (often numerous) ATP proofs found by \HH
itself. This means that for each \Flyspeck theorem, all human-written
\HOLLight proofs of all previous theorems are assumed to be known,
together with all their ATP proofs found already by \HH, but
nothing is known about the current conjecture and the following parts
of the library (they do not exist yet).

So far, systems like \HH (similar systems include \Sledgehammer/\MaSh \citep{KuhlweinBKU13}, 
\MizAR~\citep{abs-1109-0616,KaliszykU13b} 
and MaLARea~\citep{US+08}) have only used the set of \textit{named library theorems} for
proving new conjectures and thus also for the premise-selection
learning. This is usually a reasonable set of theorems to start with,
because the human mathematicians have years of experience with
structuring the formal libraries. On the other hand, there is no
guarantee that this set is in any sense optimal, both for the human
mathematicians and for the ATPs. The following three
observations indicate that the set of human-named theorems may be suboptimal:
\begin{itemize}
\item[\textit{Proofs of different length:}] The human-named theorems may differ
  considerably in the length of their proofs. The human naming is based
  on a number of (possibly traditional/esthetical) criteria that may
  sometimes have little to do with a good structuring of the library.
\item[\textit{Duplicate and weak theorems:}] The large collaboratively-build
  libraries are hard to manually guard against duplications and naming
  of weak versions of various statements. The experiments with the
  \MoMM system over the \Mizar library~\citep{Urban06-ijait} and with the recording of the \Flyspeck library~\citep{KaliszykK13} have shown
  that there are a number of subsumed and duplicated theorems, and that some
  unnamed strong lemmas are proved over and over again.
\item[\textit{Short alternative proofs:}] The experiments with
  AI-assisted ATP over the \Mizar and \Flyspeck
  libraries~\citep{AlamaKU12,holyhammer} have shown that the combined
  AI/ATP systems may sometimes find alternative proofs that are much
  shorter and very different from the human proofs, again turning some
  ``hard'' named theorems into easy corollaries.
\end{itemize}

Suboptimal naming may obviously influence the performance of the
current large-theory systems.
If many important lemmas are
omitted by the human naming, the ATPs will have to find them over and
over when proving the conjectures that depend on such lemmas. On the
other hand, if many similar variants of one theorem are named, the
current premise-selection methods might focus too much on those
variants, and fail to select the complementary theorems that are also
necessary for proving a particular conjecture.\footnote{This behavior
  obviously depends on the premise-selection algorithm. It
  is likely to occur when the premise selection is mainly based on
  symbolic similarity of the premises to the conjecture. It is less
  likely to occur when complementary semantic selection criteria are additionally
  used as, e.g.,  in SRASS~\citep{SutcliffeP07} and MaLARea~\citep{US+08}.}

To various extent, this problem might be remedied by the
alternative learning/guidance methods (ii) and (iii) mentioned in Section~\ref{Introduction}:
 Learning of internal ATP guidance using for example
Veroff's \textit{hint technique}~\citep{Veroff96}, and learning of suitable ATP strategies
using systems like BliStr~\citep{blistr}. But these methods are so far much more
experimental in the large-theory setting than premise
selection.\footnote{In particular, several initial experiments done so
  far with Veroff's hints 
over the MPTPChallenge and
  MPTP2078 benchmarks were so far unsuccessful.} That is why we
propose and explore here the following lemma-mining approach:

\begin{enumerate}
\item Considering (efficiently) the detailed graph of all atomic inferences
  contained in the ITP libraries. Such a graph has millions of nodes
  for the core \HOLLight corpus, and hundreds of millions of nodes for the whole \Flyspeck.
\item Defining over such large proof graphs efficient criteria 
  that select a smaller set of the strongest
  and most orthogonal lemmas from the corpora.
\item Using such lemmas together with (or instead of) the human-named
  theorems for 
proving new conjectures
  over the 
corpora.
\end{enumerate}

\section{Overview of Related Work and Ideas }
\label{Related}

A number of ways how to measure the quality of lemmas and how to use
them for further reasoning have been proposed already, particularly in
the context of ATP systems and proofs.  
Below we summarize recent approaches and tools that initially seemed most relevant to our work.

Lemmas are an essential part of various ATP
algorithms. State-of-the-art ATPs such as Vampire~\citep{Vampire}, E~\citep{Sch02-AICOMM} and Prover9~\citep{McC-Prover9-URL}
implement various variants of the ANL loop~\citep{WO+84}, resulting in hundreds to
billions of lemmas inferred during the prover runs. This gave rise to
a number of efficient ATP indexing techniques, redundancy control
techniques such as subsumption, and also fast ATP heuristics (based on
weight, age, conjecture-similarity, etc.) for choosing the best lemmas
for the next inferences. Several ATP methods and tools work with
such ATP lemmas.  Veroff's \textit{hint technique}~\citep{Veroff96} extracts the best
lemmas from the proofs produced by successful Prover9 runs and uses
them for directing the proof search in Prover9 on related problems. A
similar lemma-extracting, generalizing and proof-guiding technique (called \textit{E Knowledge Base -- EKB}) was
implemented by Schulz in E prover as a part of his PhD thesis~\citep{Sch00}. 

Schulz also implemented the \epcllemma tool that estimates the
best lemmas in an arbitrary DAG (directed acyclic graph) of inferences. 
Unlike the hint-extracting/guiding 
methods, this tool works not
just on the handful of lemmas involved in the final refutational
proof, but on the typically very large number of lemmas produced
during the (possibly unfinished) ATP runs.  The \epcllemma's criteria
for selecting the next best lemma from the inference DAG are: (i) the size of
the lemma's inference subgraph based at the nodes that are either axioms
or already chosen (better) lemmas, and (ii) the weight of the
lemma. This lemma-selection process may be run recursively, until a stopping
criterion (minimal lemma quality, required number of lemmas, etc.) is
reached. Our algorithm for selecting \HOLLight lemmas (Section~\ref{Good}) is
quite similar to this.

\AGIntRater~\citep{PuzisGS06} is a tool that computes various characteristics of the
lemmas that are part of the final refutational ATP proof and aggregates them into
an overall \emph{interestingness} rating. These characteristics include:
obviousness, complexity, intensity, surprisingness, adaptivity, focus,
weight, and usefulness, see~\citep{PuzisGS06} for details. \AGIntRater so
far was not directly usable on our data for various reasons
(particularly the size of our graph), but we might re-use and try to
efficiently implement some of its ideas later.

\citet{Pud06-ESCoR} has conducted experiments over several datasets
with automated re-use of lemmas from many existing ATP proofs in order to find
smaller proofs and also to attack unsolved problems. This is similar
to the hints technique, however more automated and closer to our large-theory setting
(hints have so far been successfully applied mainly in small algebraic
domains). To interreduce the large number of such lemmas with respect
to subsumption he used the \textit{CSSCPA}~\citep{Sut01-LPAR} subsumption tool based on the E prover by Schulz
and Sutcliffe. \MoMM~\citep{Urban06-ijait} adds a number of large-theory features to
CSSCPA. It was used for (i) fast interreduction of million of lemmas extracted (generalized)
from the proofs in the \Mizar library, and (ii) as an early ATP-for-ITP hammer-style tool for
completing proofs in \Mizar with the help of the whole \Mizar library. All
library lemmas can be loaded, indexed and considered for each query, 
however the price for this breadth of coverage is
that the inference process is limited to subsumption extended with
\Mizar-style dependent types. 

\AGIntRater and \epcllemma use a lemma's position in the inference graph
as one of the lemma's characteristics that contribute to its
importance. There are also purely graph-based algorithms that try to
estimate a relative importance of nodes in a graph. In particular,
research of large graphs became popular with the appearance of the
World Wide Web and social networks. Algorithms such as
\textit{PageRank}~\citep{page98pagerankTechreport} (eigenvector
centrality) have today fast approximative implementations that easily
scale to billions of nodes.

\section{The Proof Data}
\label{Data}

We consider two corpora: the core \HOLLight corpus (SVN version 179) and the
\Flyspeck corpus (SVN version 3437). 
The core \HOLLight corpus contains
of 2,239 named theorems, while the \Flyspeck corpus consists of 23,323 named theorems.
The first prerequisite for implementing and running interesting lemma-finding algorithm is the 
extraction of the full dependency graph containing all intermediate steps (lemmas),
 and identification of the named top-level theorems among them.

There are three
issues with the named theorems that we initially need to address. First, many theorems in \HOLLight are
conjunctions. It is often the case
that lemmas that deal with the same constant or theory are put in the same theorem,
so that they can be passed to tactics and decision procedures as a single argument
rather than a list. Second, a single theorem may be given multiple names. This is
especially common in case of larger formalizations like \Flyspeck. Third, even if
theorems are not syntactically equal they may be alpha equal. \HOLLight does not
natively use de Bruijn indices for representing variables, i.e., two alpha-convertible
versions of the same theorems will be kept in the proof trace if they differ in
variable names. Therefore the first operation we perform is to find a unique name for each separate
top-level conjunct. The data sizes and processing times of this first phase can be
found in Table~\ref{t:topthms}.

\begin{table}[htb]\centering
\begin{tabular}{ccc}\toprule
& \HOLLight (179) & \Flyspeck (3437) \\\midrule
Named theorems & 2,239 & 23,323\\
Distinct named conjuncts & 2,542 & 24,745\\
Constant definitions & 234&2,106 \\
Type definitions & 18& 29\\
Processing time & 2m09s & 327m56s\\
Processing memory & 214MB & 1,645MB\\
\bottomrule\end{tabular}
\caption{\label{t:topthms} The top-level available data and processing statistics of the analyzed corpora.}
\end{table}

We next look at all the available intermediate lemmas,
each of them corresponding to one of the LCF-style kernel inferences done by \HOLLight. 
The number of these lemmas
when processing \Flyspeck
is around 1.7 billion. 
Here, already performing the above mentioned reduction is hard since the whole graph with the 1.7 billion \HOLLight formulas 
can be considered big data: it fits
neither in memory nor on a single hard disk. Therefore we 
perform the first graph reductions already when recording the proof trace.

To obtain the full inference graph for \Flyspeck we run the
proof-recording version of \HOLLight~\citep{KaliszykK13} patched to
additionally remember all the intermediate lemmas. Obtaining such
trace for \Flyspeck takes 29 hours of CPU time and 56 GB of RAM on an
AMD Opteron 6174 2.2 GHz
Because of the memory consumption we initially consider
two versions: a) de-duplicating all the intermediate lemmas within
a named theorem; we call the graph obtained in this way \texttt{TRACE0}, and b)
de-duplicating all the lemmas; which we call \texttt{TRACE1}. The sizes
of the traces are presented in Table~\ref{t:traces}. This time and memory consumption are much
lower when working only with the core \HOLLight, where a further graph
optimization in this step could already be possible.

\begin{table}[htb]\centering
\begin{tabular}{crrrr}\toprule
& \multicolumn{2}{c}{\HOLLight graph} & \multicolumn{2}{c}{\Flyspeck graph}  \\
 \cmidrule(lr){2-3}  \cmidrule(lr){4-5}
& \multicolumn{1}{c}{nodes} & \multicolumn{1}{c}{edges} &
  \multicolumn{1}{c}{nodes} & \multicolumn{1}{c}{edges} \\\midrule
kernel inferences & 8,919,976 & 10,331,922& 1,728,861,441 & 1,953,406,411\\
\texttt{TRACE0}  & 2,435,875 & 3,476,767 & 206,699,009 & 302,799,816\\
\texttt{TRACE1} 
& 2,076,682 & 3,002,990 & 159,102,636 & 233,488,673 \\
tactical inferences & 148,514 & 594,056 & 11,824,052 & 42,296,208\\
tactical trace   & 22,284 & 89,981 & 1,067,107 & 4,268,428\\
\bottomrule\end{tabular}
\caption{\label{t:traces} The sizes of the inference graphs.}
\end{table}

There are 1,953,406,411 inference edges between the unique \Flyspeck lemmas.
During the proof recording we additionally
export the information about the symbol weight (size) of each lemma,
and for the small \HOLLight traces also the lemma's normalized form that serially numbers bound and free variables
and tags them with their types. This information is later used for
external postprocessing, together with the information about which theorems where originally named. 
The initial segment of the \Flyspeck proof trace is presented in
Fig.~\ref{fig:trace}, all the traces are available online.\footnote{\url{http://cl-informatik.uibk.ac.at/~cek/lemma_mining/}}

\begin{figure}[htb]
\begin{holnb}\small\holthm{
F13        #1, Definition (size 13): T <=> (\Lam{}A0. A0) = (\Lam{}A0. A0)
R9         #2, Reflexivity (size 9): (\Lam{}A0. A0) = (\Lam{}A0. A0)
R5         #3, Reflexivity (size 5): T <=> T
R5         #4, Reflexivity (size 5): (<=>) = (<=>)
C17 4 1    #5, Application(4,1):     (<=>) T = (<=>) ((\Lam{}A0. A0) = (\Lam{}A0. A0))
C21 5 3    #6, Application(5,3):     (T <=> T) <=> (\Lam{}A0. A0) = (\Lam{}A0. A0) <=> T
E13 6 3    #7, EQ_MP(6,3) (size 13): (\Lam{}A0. A0) = (\Lam{}A0. A0) <=> T
}\end{holnb}
\caption{\label{fig:trace}Initial segment of the \HOLLight theorem trace commented with the numbers of the steps and the theorems derived by the steps.}
\end{figure}

\subsection{Initial Post-processing and Optimization of the Inference Traces}

During the proof recording, only exact duplicates are easy to
detect. As already explained in the previous Section, \HOLLight does
not natively use de Bruijn indices for representing variables, so the
trace may still contain alpha-convertible versions of the
same theorems. Checking for alpha convertibility during the proof
recording would be possible, however is not obvious since in the \HOLLight's LCF-style
approach alpha conversion itself results in multiple kernel inferences.
In order to avoid performing term-level renamings we keep the original proof trace untouched, and implement its
further optimizations as external postprocessing of the trace.

In particular, to merge alpha convertible lemmas in a proof trace $T$,
we just use the above mentioned normalized-variable representation of
the lemmas as an input to an external program that produces a new
version of the proof trace $T'$.  This program goes through the trace $T$
and replaces references to each lemma by a reference to the earliest
lemma in $T$ with the same normalized-variable representation. The proofs of
the later named alpha variants of the lemmas in $T$ are however still kept in the new
trace $T'$, because such proofs are important when computing the
usage and dependency statistics over the normalized lemmas.  We
have done this postprocessing only for the core \HOLLight
lemmas,
because printing out of the
variable-normalized version of the 150,142,900 partially de-duplicated \Flyspeck lemmas 
is currently not feasible on our hardware. 
From the 2,076,682 partially de-duplicated core \HOLLight lemmas
1,076,995 are left after this stronger normalization. We call such further post-processed
graph \texttt{TRACE2}.

It is clear that such post-processing operations can be
implemented in various ways. In this case, some original
information about the proof graph is lost, while some information
(proofs of duplicate lemmas) is still kept, even though it could be also
pruned from the graph, producing a differently normalized
version.

\subsection{Obtaining Shorter Traces from the Tactic Calls}
Considering the \HOL kernel proof steps as the atomic steps in construction
of intermediate lemmas has (at least) three drawbacks. First, the pure size of
the proof traces makes it hard to scale the lemma-mining procedures to big developments like \Flyspeck.
Second, the multitude of steps that arise when applying simple \HOLLight decision procedures
overshadows the interesting parts of the proofs. It is not uncommon for a simple
operation, like a normalization of a polynomial, to produce tens of thousands of core kernel
inferences. Third, some operations (most notably the \HOLLight simplifier) produce kernel
inferences in the process of proof search. Such inferences are not only
uninteresting (as in the previous case), but often useless for the final proof.

In order to overcome the above three issues encountered in the first experiments, we followed by gathering data at the level
of the \HOLLight \emph{tactic} steps~\citep{Harrison96}. The execution of each \HOLLight tactic produces a new goal state together with
a justification function that produces an intermediate lemma. In this approach, instead of considering all kernel steps, we will consider
only the lemmas produced by the justification functions of tactics. The \HOLLight tactics work
on different levels. The tactics
executed by the user and visible in the proof script form the outermost layer. However
most of the tactics are implemented as \OCaml functions that inspect the
goal and execute other (smaller) tactics. If we unfold such internal executions of tactics
recursively, the
steps performed are of a similar level of detail as in typical natural deduction proofs.

This could give us a trace that is slightly smaller than the typical trace of the kernel
inferences; however the size is still of the same order of magnitude. In order to efficiently
process large formal developments we decided to look at an intermediate level: only at
the tactics that are composed using \emph{tactic combinators}~\citep{Harrison96}.

In order to patch the tactic combinators present in \HOLLight and \Flyspeck
it is enough to patch the three building blocks of tactic combinators: \texttt{THEN},
\texttt{THENL}, and \texttt{by}. Loading \Flyspeck with these functions patched
takes about 25\% more time than the original and requires 6GB of memory to remember
all the 20 million new intermediate theorems. This is significantly less than the
patched kernel version and the produced graph can be reasonably optimized.

The optimizations performed on the level of named theorems can be done
here again: recursively splitting conjunctions and normalizing the quantifiers, as well
as the premises we get 2,014,505 distinct conjuncts. After alpha-normalization this leaves
a trace with 1,067,107 potential intermediate lemmas. In order to find dependencies between
the potential intermediate lemmas we follow the approach by~\citet{KaliszykK13} which needs
a second dependency recording pass over the whole \Flyspeck.

The post-processed tactics dependency graph
has 4,268,428 edges and only 2,145 nodes have no dependencies. The comparison of all
the traces can be seen in Table~\ref{t:traces}. The data is written in the same format
as the \HOL kernel inference data, so that we can use the same predictors. An excerpt
from the tactical trace coming from the proof of \texttt{MAP_APPEND} is presented in
Fig.~\ref{fig:tracetac}.

\begin{figure}[htb]
\begin{holnb}\normalsize\holthm{
X29 3377 3371       #3437, Rewriting with two given theorems, size 29:
          |- !l2. MAP f (APPEND [] l2) = APPEND (MAP f []) (MAP f l2)

X66 3378 3372       #3438, Rewriting with two given theorems, size 66:
          |- !t. (!l2. MAP f (APPEND t l2) = APPEND (MAP f t) (MAP f l2))
                           ==> (!l2. MAP f (APPEND (CONS h t) l2) =
                               APPEND (MAP f (CONS h t)) (MAP f l2))

X33 3321 3437 3438  #3439, List induction, size 33:
          |- !f l1 l2. MAP f (APPEND l1 l2) = APPEND (MAP f l1) (MAP f l2)
}\end{holnb}
\caption{\label{fig:tracetac}An excerpt of the tactical trace showing the dependencies
between the goal states in the proof of \texttt{MAP_APPEND}. For simplicity we chose an
excerpt that shows the theorems created by the direct application of a tactic that does
not call other tactics (\texttt{LIST_INDUCT_TAC}). This means that all the theorems
created in this part of the trace directly correspond to goals visible to the proof-assistant
user.}
\end{figure}

\subsection{Other Possible Optimizations}
The ATP experiments described below use only the four versions
of the proof trace (\texttt{TRACE0}, \texttt{TRACE1}, \texttt{TRACE2}, 
and the tactical trace) described above, but we have also explored some
other normalizations. A particularly interesting optimization from the
ATP point of view is the removal of subsumed lemmas. An initial
measurement with the (slightly modified) \MoMM system done on the
clausified first-order versions of about 200,000 core \HOLLight lemmas has
shown that about 33\% of the clauses generated from the lemmas are
subsumed. But again, ATP operations like subsumption interact with the
level of inferences recorded by the \HOLLight kernel in nontrivial
ways. It is an interesting task to define exactly how the original
proof graph should be transformed with respect to such operations, and
how to perform such proof graph transformations efficiently over the whole
\Flyspeck.

\section{Selecting Good Lemmas}
\label{Good}

Several approaches to defining the notion of a useful/interesting
lemma are mentioned in Section~\ref{Related}. There are a number of
ideas that can be explored and combined together in various ways, but
the more complex methods (such as those used by \AGIntRater) are not
yet directly usable on the large ITP datasets that we have. So far, we
have experimented mainly with the following techniques:

\begin{enumerate}
\item A direct \OCaml implementation of lemma quality metrics based on the \HOLLight
   proof-recording data structures. 
\item Schulz's \epcllemma and its minor modifications.
\item PageRank, applied in various ways to the proof trace.
\item Graph cutting algorithms with modified weighting function.
\end{enumerate}

\subsection{Direct Computation of Lemma Quality}\label{sec:qdirect}
The advantage of the direct \OCaml implementation is that no export to external tools is
necessary and all the information collected about the lemmas by the
\HOLLight proof recording is directly available.  The basic factors
that we use so far for defining the quality of a lemma $i$ are its:
(i) set of direct proof dependencies $d(i)$ given by the proof trace,
(ii) number of recursive dependencies $D(i)$, (iii) number of
recursive uses $U(i)$, and (iv) number of HOL symbols (HOL weight)
$S(i)$. When recursively defining $U(i)$ and $D(i)$ we assume that in
general some lemmas may already be named ($k \in Named$) and some
lemmas are just axioms ($k \in Axioms$). Note that in \HOLLight there
are many lemmas that have no dependencies, but formally they are still
derived using for example the reflexivity inference rule (i.e., we do
not count them among the \HOLLight axioms). The recursion when defining $D$ thus
stops at axioms, named lemmas, and lemmas with no dependencies. The
recursion when defining $U$ stops at named lemmas and unused lemmas. Formally:

\begin{defn}[Recursive dependencies and uses]
\begin{align*}
D(i) &=
\begin{cases}
1 & \text{if $i \in Named \lor i \in Axioms$},\\
\sum\limits_{j \in d(i)} D(j) & \text{otherwise.}\\
\end{cases}
\\
U(i) &=
\begin{cases}
1 & \text{if $i \in Named$,}\\
\sum\limits_{i \in d(j)} U(j) & \text{otherwise.}
\end{cases}
\end{align*}
\end{defn}
In particular, this means that 
$$D(i) = 0 \iff d(i) = \emptyset \land \lnot (i \in Axioms)$$ 
and also that
$$U(i) = 0 \iff \forall j \lnot (i \in d(j))$$
These basic characteristics are combined into the following lemma
quality metrics $Q_1(i)$, $Q_2(i)$, and $Q_3(i)$. $Q_1^r(i)$ is a
generalized version of $Q_1(i)$, which we (apart from $Q_1$) test for $r \in \{0, 0.5, 1.5, 2 \}$:

\begin{defn}[Lemma quality]
\begin{equation*}
\begin{aligned}
Q_1(i) &= \frac{U(i) * D(i)}{S(i)}\\
Q_2(i) &= \frac{U(i) * D(i)}{S(i)^2}\\
\end{aligned}
\qquad
\begin{aligned}
Q_1^r(i) &= \frac{U(i)^r * D(i)^{2-r}}{S(i)}\\
Q_3(i) &= \frac{U(i) * D(i)}{1.1^{S(i)}}\\
\end{aligned}
\end{equation*}
\end{defn}
The justification behind these definitions are the following heuristics: 
\begin{enumerate}
\item The higher is $D(i)$, the more necessary it is to remember the lemma $i$, because it will be harder to infer with an ATP when needed.
\item The higher is $U(i)$, the more useful the lemma $i$ is for proving other desired conjectures.
\item The higher is $S(i)$, the more complicated the lemma $i$ is in
  comparison to other lemmas. In particular, doubled size may often
  mean in \HOLLight that $i$ is just a conjunction of two other
  lemmas.\footnote{The possibility to create conjunctions is quite a
    significant difference to the clausal setting handled by
    the existing tools. A longer clause is typically weaker, while
    longer conjunctions are stronger. A dependence on a longer
    conjunction should ideally be treated by the evaluating heuristics
    as a dependence on the multiple conjuncts. Note that for the tactical 
trace we already split all conjunctions in the trace.}
\end{enumerate}

\subsection{Lemma Quality via \epcllemma}\label{sec:qepcl}

Lemma quality in \epcllemma is defined on clause inferences
recorded using E's native PCL protocol. The lemma quality computation
also takes into account the lemmas that have been already named, and with minor implementational variations it can be expressed using $D$ and $S$ as follows:
\begin{equation*}
EQ_1(i) = \frac{D(i)}{S(i)}\\
\end{equation*}
The difference to $Q_1(i)$ is that $U(i)$ is not used, i.e., only the
cumulative effort needed to prove the lemma counts, together with its
size (this is also very close to $Q_1^r(i)$ with $r = 0$).  The main advantage of using \epcllemma is its fast and robust
implementation using the E code base. This allowed us to load in
reasonable time (about one hour) the whole \Flyspeck proof trace into
\epcllemma, taking 67 GB of RAM.  Unfortunately, this experiment showed
that \epcllemma assumes that $D$ is always an integer. This is likely
not a problem for \epcllemma's typical use, but on the \Flyspeck graph
this quickly leads to integer overflows and wrong results. To a
smaller extent this shows already on the core \HOLLight proof
graph. A simple way how to prevent the overflows was to modify \epcllemma to use instead of $D$ the longest chain of inferences $L$:
\begin{equation*}
L(i) =
\begin{cases}
1 & \text{if $i \in Named \lor i \in Axioms$},\\
max_{j \in d(i)} (1+L(j)) & \text{otherwise.}\\
\end{cases}
\end{equation*}
This leads to:
\begin{equation*}
EQ_2(i) = \frac{L(i)}{S(i)}\\
\end{equation*}
Apart from this modification, only minor changes were needed to make
\epcllemma work on the \HOLLight data. The proof trace was
expressed as a PCL proof (renaming the HOL inferences into E
inferences), and TPTP clauses were used instead of the original \HOL clauses.
We additionally compared two strategies of creating the TPTP clauses.
First we applied the \MESON translation to the \HOL clause,
second we tried to create artificial TPTP clauses of the size
corresponding to the size of the \HOL clause.

\subsection{Lemma Quality via PageRank}\label{sec:qpagerank}
PageRank (eigenvector centrality of a graph) is a method that assigns
weights to the nodes in an arbitrary directed graph (not just DAG)
based on the weights of the neighboring nodes (``incoming links''). In
more detail, the weights are computed as the dominant eigenvector of
the following set of equations:
\begin{equation*}
PR_1(i) = \frac{1 - f}{N} + f \sum\limits_{i \in d(j)}\frac{PR_1(j)}{|d(j)|}\\
\end{equation*}
where $N$ is the total number of nodes and $f$ is a damping factor,
typically set to 0.85. The advantage of using PageRank is that there
are fast approximative implementations that can process the whole
\Flyspeck proof graph in about 10 minutes using about 21 GB RAM, and
the weights of all nodes are computed simultaneously in this time.

This is however also a disadvantage in comparison to the previous
algorithms: PageRank does not take into account the lemmas that have
already been selected (named). The closer a lemma $i$ is to an
important lemma $j$, the more important $i$ will be. Modifications
that use the initial PageRank scores for more advanced clustering
exist~\citep{AvrachenkovDNPS08} and perhaps could be used to mitigate
this problem while still keeping the overall processing reasonably
fast. Another disadvantage of PageRank is its ignorance of the lemma
size, which results in greater weights for the large conjunctions that
are used quite often in \HOLLight. $PR_2$ tries to counter that:
\begin{equation*}
PR_2(i) = \frac{PR_1(i)}{S(i)}\\
\end{equation*}
$PR_1$ and $PR_2$ are based on the idea that a lemma is important if
it is needed to prove many other important lemmas. This can be again
turned around: we can define that a lemma is important if it
depends on many important lemmas. This is equivalent to computing the
reverse PageRank and its size-normalized version:
\begin{equation*}
PR_3(i) = \frac{1 - f}{N} + f \sum\limits_{i \in u(j)}\frac{PR_3(j)}{|u(j)|}\\
\qquad
PR_4(i) = \frac{PR_3(i)}{S(i)}\\
\end{equation*}
where $u(j)$ are the direct uses of the lemma $j$, i.e., $i \in u(j) \iff j \in d(i)$. The two ideas can again be combined (note that the sum of the PageRanks of all nodes is always 1):
\begin{equation*}
PR_5(i) = {PR_1(i) + PR_3(i)} \qquad PR_6(i) = \frac{PR_1(i) + PR_3(i)}{S(i)}\\
\end{equation*}

\subsection{Lemma Quality using  Graph Cut}\label{sec:qcut}

The approaches so far tried to define what a ``good'' lemma is using
our intuitions coming from mathematics. Here we will try to estimate
the impact that choosing certain lemmas will have on the final dependency
graph used for the learning framework.

Choosing a subset of the potential intermediate lemmas can be considered
a variant of the graph-theoretic problems of finding a cut with certain
properties. We will consider only cuts that respect the chronological
order of theorems in the library. Since many of the graph-cut algorithms
(for example maximum cut) are NP-complete, we decide to build the cut
greedily adding nodes to the cut one by one.

\begin{figure}[htb]
\centering
\includegraphics[width=.6\textwidth]{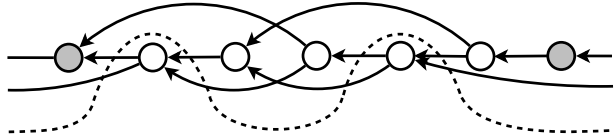}
\caption{\label{fig:maxcut} An example cut of the dependency graph that
  respects the chronological order of the library. The already named theorems
  are marked in gray.}
\end{figure}

Given a graph where certain nodes are already named (marked gray in
the Figure~\ref{fig:maxcut}) we want to estimate the impact of choosing
a new lemma on the evaluation. In the evaluation, we will compute the
dependency graph of all the gray nodes together with the newly chosen
one. The final graph represents the human dependencies, which means that
theorems are ATP-provable using exactly these dependencies. By minimizing
the number of edges in this final graph we make the average number of premises
in the problems smaller which should make the problems easier to prove.
The assumption here, is that training our estimators on theorems that are
easier to prove makes the resulting AI/ATP system stronger.

In order to minimize the number of edges in the final graph we will investigate
what is the impact of adding a node $n$ to the cut. We consider all the dependency
paths starting which start at $n$. On each path we select the first already named
node. All the nodes that have been selected are the dependencies that $n$ will
have in the final dependency graph. Lets denote this set as $D(n)$.
Similarly we can find all the nodes that will have $n$ as a dependency
This can be done in a similar way, taking all the paths the opposite direction
again choosing the first gray node on each path. Lets denote the nodes
found as $U(n)$. These nodes will have $n$ as a dependency if $n$ is chosen to
be in the cut.

\begin{thm}
Adding a node $n$ to the cut $c$ will decrease the number of edges in the final
graph by $|D(n)| * |U(n)| - |D(n)| - |U(n)|$.
\end{thm}
\begin{pf}
  With the cut $c$ the edges in the final graph include all the edges between
  the nodes in $D(n)$ and $U(n)$. Adding the node $n$ to $c$ these $|D(n)| * |U(n)|$
  edges will be replaced by the dependencies from each element of $U(n)$ to $n$
  ($|U(n)|$ many of them) and the dependencies from $n$ to all the elements of
  $D(n)$ ($|D(n)|$ many of them).
\end{pf}

The algorithm manipulates sets of nodes rather than numbers, which makes it
significantly slower than all the previously described ones. We will test this
algorithm only for up to 10,000 lemmas as already finding them takes 11 CPU hours.
Similarly to the algorithms in the previous subsections we try to
limit the effect of large theorems on the algorithm by considering
also the size normalized version:

\begin{equation*}
MC_1(i) = |D(i) * U(i)| - |D(i)| - |D(i)| \qquad MC_2(i) = \frac{MC_1(i)}{S(i)}
\end{equation*}

\subsection{Selecting Many Lemmas}

From the methods described above, only the various variants of
PageRank ($PR_i$) produce the final ranking of all lemmas in one
run. Both \epcllemma ($EQ_i$) and our custom methods ($Q_i$, $MC_i$) are
parametrized by the set of lemmas ($Named$) that have already been
named. When the task is to choose a predefined number of the best
lemmas, this naturally leads to the recursive lemma-selection Algorithm~\ref{a:manylemmas} (used also by \epcllemma).
\begin{algorithm}[!ht]
\caption{Best lemmas\label{a:manylemmas}}
\begin{algorithmic}[1]
\Require{a lemma-quality metric $Q$, set of lemmas $Lemmas$, an initial set of named lemmas $Named_0 \subset Lemmas$, and a required number of lemmas $M$}
\Ensure{set $Named$ of $M$ best lemmas according to $Q$}
\State $Named \gets Named_0$
\State $m \gets 0$
  \While{$m < M$}
    \For{$i \in Lemmas$}
      \State \Call{Calculate}{$Q_{Named}(i)$}
    \EndFor
    \State $j \gets argmax \{ Q_{Named}(i) : i \in Lemmas \setminus Named \}$
    \State $Named \gets Named \cup \{j\}$
    \State $m \gets m+1$
  \EndWhile
  \State \Call{Return}{$Named$}
\end{algorithmic}
\end{algorithm}

There are two possible choices of the initial set of named lemmas $Named_0$ in Algorithm~\ref{a:manylemmas}: either the empty set, or
the set of all human-named theorems. This choice depends on whether we
want to re-organize the library from scratch, or whether we just want to
select good lemmas that complement the human-named theorems. Below we
experiment with both approaches. Note that this algorithm is currently
quite expensive: the fast \epcllemma implementation takes 65 seconds to
update the lemma qualities over the whole \Flyspeck graph after each
change of the $Named$ set. This means that with the kernel-based inference trace
(\texttt{TRACE1})
producing the first 10,000 \Flyspeck lemmas takes 180 CPU hours.
That is why most of the experiments are limited to the core \HOLLight
graph and \Flyspeck tactical graph where this
takes about 1 second and 3 hours respectively.

\section{Evaluation Scenarios and Issues}
\label{Scenarios}
To assess and develop the lemma-mining methods we define several
evaluation scenarios that vary in speed, informativeness and
rigor. The simplest and least rigorous is the \textit{expert-evaluation} scenario:
We can use our knowledge of the formal corpora to quickly see if the
top-ranked lemmas produced by a particular method look
plausible.

The \textit{cheating ATP} scenario uses the full proof graph of a
corpus to compute the set of the (typically 10,000) best lemmas
($BestLemmas$) for the whole corpus. Then the set of newly named
theorems ($NewThms$) is defined as the union of $BestLemmas$ with the
set of originally named theorems ($OrigThms$): $NewThms \coloneqq
BestLemmas \cup OrigThms$. The derived graph $G_{NewThms}$ of direct
dependencies among the elements of $NewThms$ is used for ATP
evaluation, which may be done in two ways: with human selection and
with AI selection. When using human selection, we try to prove each
lemma from its parents in $G_{NewThms}$. When using AI selection, we
use the chronological order (see Section~\ref{Approach}) of $NewThms$
to incrementally train and evaluate the $k$-NN machine learner~\citep{EasyChair:74} on the
direct dependencies from $G_{NewThms}$. This produces for each new
theorem an ATP problem with premises advised by the learner trained on
the $G_{NewThms}$ dependencies of the preceding new theorems. This
scenario may do a lot of cheating, because when measuring the ATP
success on $OrigThms$, a particular theorem $i$ might be proved with
the use of lemmas from $NewThms$ that have been stated for the first time
only in the original proof of $i$ (we call such lemmas
\textit{directly preceding}). In other words, such lemmas did not
exist before the original proof of $i$ was started, so they could not
possibly be suggested by lemma-quality metrics for proving $i$. Such
directly preceding lemmas could also be very close to $i$, and thus
equally hard to prove.

The \textit{almost-honest ATP} scenario does not allow the use of the directly
preceding new lemmas. The dependencies of each $i \in NewThms$ may consist only of the previous $OrigThms$
and the lemmas that precede them. Directly preceding new lemmas are replaced by
their closest $OrigThms$ ancestors. This scenario is still not fully honest,
because the lemmas are computed according to their lemma quality
measured on the full proof graph. In particular, when proving an early
theorem $i$ from $OrigThms$, the newly used parents of $i$ are lemmas
whose quality was clear only after taking into account the theorems
that were proved later than $i$. These theorems and their proofs
however did not exist at the time of proving $i$. Still, we consider
this scenario sufficiently honest for most of the ATP evaluations done with
the whole core \HOLLight dataset and the representative subset of the \Flyspeck
dataset.

The \textit{fully-honest ATP} scenario removes this last objection, at
the price of using considerably more resources for a single evaluation. For
each originally named theorem $j$ we limit the proof graph used for
computing $BestLemmas$ to the proofs that preceded $j$. Since
computing $BestLemmas$ for the whole core \HOLLight takes at least
three hours for the $Q_i$ and $EQ_i$ methods, the full evaluation on
all 1,954 core \HOLLight theorems would take about 2,000 CPU
hours. That is why we further scale down this evaluation by doing it
only for every tenth theorem in core \HOLLight.

The \textit{chained-conjecturing ATP} scenario is similar to the
cheating scenario, but with limits imposed on the directly preceding
lemmas. In $chain_1$\textit{-conjecturing}, any (possibly directly
preceding) lemma used to prove a theorem $i$ must itself have an ATP
proof using only $OrigThms$. In other words, it is allowed to guess
good lemmas that still do not exist, but such lemmas must not be hard
to prove from $OrigThms$. Analogously for
$chain_2$\textit{-conjecturing} (resp. $chain_N$), where lemmas
provable from $chain_1$-lemmas (resp. $chain_{N-1}$) are allowed to be
guessed.  To some extent, this scenario measures the theoretical ATP
improvement obtainable with guessing of good intermediate lemmas.

\section{Experiments}
\label{Experiments}

In total, we have performed experiments with 180 different strategies for adding
new lemmas based on the kernel inference traces, and with 164 different
strategies for adding new lemmas based on the tactical traces.
The ATP experiments are done on the same hardware and using the same
setup that was used for the earlier evaluations described
in~\citep{holyhammer,EasyChair:74}: All ATP systems are run with 30s time
limit on a 48-core server with AMD Opteron 6174 2.2 GHz CPUs, 320 GB
RAM, and 0.5 MB L2 cache per CPU. 

In order to find the exact \HOL formulas corresponding to the new
lemmas (known only as nodes in a graph) coming from mining the kernel
inference traces, we first have to process the formalization again
with a patched kernel that takes the lemma numbers as a parameter and
exports also the statements of the selected new lemmas.
This is no longer necessary for the tactic data, since the formula
statements can be stored together with the proof graph during the
first run.
The slowest part of our setup is computing the formula features needed
for the machine learning. For the experiments with the 
kernel inference lemmas, the features of each final set of selected lemmas ($NewThms$) 
are computed independently, since we cannot pre-compute the features of all the lemmas in the kernel traces.
In case of the \Flyspeck tactical trace we can directly compute the
features of all of the over 1 million lemmas. Due to their size (the intermediate lemmas are often large implications), 
it takes 28 hours to extract and normalize~\citep{holyhammer} all the features.
The sum of the counts of such features over all these lemmas is 63,433,070, but there are just 383,304 unique features in these lemmas.
Even for the extreme case of directly using and predicting premises for
 all the lemmas from the \Flyspeck tactical trace without any preselection, 
our k-NN predictor can perform all the one million predictions in about 30 hours, taking
0.11s per prediction. Predictions are translated from the \HOL logic into FOF problems~\citep{holyhammer} and ATPs are
run on them in the usual way to make the evaluations.

In order to compare the new results with the extensive experimental
results obtained over the previous versions of \HOLLight and \Flyspeck
used in~\citep{holyhammer}, we first detect the set of theorems that
are preserved between the different versions. This is done by using
the recursive content-based naming of symbols and theorems that we
have developed for re-using as much information between different
library versions in the \HH online service~\citet{hhmcs}.
In case of \HOLLight the complete
set of 1954 core \HOLLight theorems evaluated in previous evaluations of \HH
has been preserved, only some of the names have been changed. In case of
\Flyspeck a smaller set of 10,779 theorems is preserved. In order to perform more
experiments we further reduced the size of this set by choosing only every sixth
theorem and evaluating the performance on the resulting 1796 theorems.

\subsection{Evaluation on Core \HOLLight}

When using only the original
theorems, the success rate of the 14 most complementary AI/ATP methods developed in~\citep{holyhammer}
run with 30s time limit each and restricted to the 1954 core \HOLLight
theorems is 63.1\% (1232 theorems) and the union of all those methods solved
65.4\% (1278 theorems). 
In the very optimistic \textit{cheating} scenario (limited only to the
$Q_i$ metrics), these numbers go up to 76.5\% (1496 theorems)
resp. 77.9\% (1523 theorems). As mentioned in Section~\ref{Scenarios},
many proofs in this scenario may however be too simple because a close
directly preceding lemma was used by the
lemma-mining/machine-learning/ATP stack. This became easy to see already
when using the \textit{almost-honest} scenario, where the 14 best
methods (including also $EQ_i$ and $PR_i$) solve together only 66.2\%
(1293 theorems) and the union of all methods solves 68.9\% (1347
theorems). The performance of the various (almost-honest) new lemma-based methods is shown in Table~\ref{t:methods},
together with their comparison and combination with the old experiments.

\begin{table}[htb]\centering
\begin{tabular}{cccc}\toprule
Strategy & Theorems (\%) & Unique & Theorems \\\midrule
 $Q_{1..3}$ (direct quality, sec.~\ref{sec:qdirect}) &  62.897 & 68 &  1229 \\
 $PR_{1..5}$ (PageRank, sec.~\ref{sec:qpagerank})  &  58.700 & 17 &  1147\\
 $EQ_{1..2}$ (\epcllemma, sec.~\ref{sec:qepcl}) &  57.011 & 4  &  1114\\
 $MC_{1..2}$ (graph cut, sec.~\ref{sec:qcut})  &  47.288 & 1  &  924\\
   total   &  64.125 &    &  1253\\\midrule
 only named &  54.452 & 0  &  1064\\
   total   &  64.125 &    &  1253\\\midrule
 \HH (14 best) &  63.050 & 92 &  1232\\
 combined 14 best & 66.172 &  & 1293\\
   total   &  68.833 &    &  1345\\\bottomrule
 \end{tabular}
\caption{\label{t:methods}Comparison of the methods evaluated on the kernel traces
on the 1954 \HOLLight theorems}
\end{table}

\begin{table}[htb]\centering
\begin{tabular}{cccc}\toprule
Added theorems & Success rate & Unique & Thms\\\midrule
 \texttt{TRACE2} & 62.078 & 48 & 1213\\
 \texttt{TRACE0} & 59.365 & 12 & 1160\\
 \texttt{TRACE1} & 58.802 & 17 & 1149\\\midrule
 10,000  &63.562 & 138 &1242\\
  1,000  &55.374 & 9   &1082\\\bottomrule
\end{tabular}
\caption{\label{t:traceeval}Success rate depending on kind of trace used
  and depending on the number of added theorems}
\end{table}

The majority of the new solved problems come from the alpha-normalized
\texttt{TRACE2}, however the non-alpha normalized versions with and without
duplicates do contribute as well. When it comes to the number of theorems
added, adding more theorems seems to help significantly, see Table~\ref{t:traceeval}. We do not try
to add more than 10,000 theorems for core \HOLLight, as this is already
much bigger than the size of the library. We will add up to
one million theorems when looking at the whole \Flyspeck in the next
subsection.

For each of the strategies the success rates again depend on the
different arguments that the strategy supports. In case of direct lemma
computation considering $Q_1$ seems to give the best results, followed by $Q_2$
and $Q_3^{1.1}$; see Table~\ref{t:seqeval}. This suggest that focusing on either
$U$ or $D$ is worse than looking at the combination. For core \HOLLight size
seems not to be an issue and dividing by size gives us best results. This will
change in \Flyspeck where the real arithmetic decision procedures produce much
bigger intermediate lemmas.

\renewcommand{\arraystretch}{1.5}
\begin{table}[htb]\centering
\begin{tabular}{cccc}\toprule
Lemma quality & Success rate & Unique & Thms\\\midrule
   $Q_1$ ($\frac{U(i) * D(i)}{S(i)}$) & 58.751 & 21   & 1148\\
   $Q_2$ ($\frac{U(i) * D(i)}{S(i)^2}$)        & 57.932 & 10   & 1132\\
   $Q_3^{1.1}$ ($\frac{U(i) * D(i)}{1.1^{S(i)}}$)   & 57.523 & 8    & 1124\\
   $Q_3^{1.25}$ ($\frac{U(i) * D(i)}{1.25^{S(i)}}$)  & 53.685 & 2    & 1049\\
   $Q_3^{1.05}$ ($\frac{U(i) * D(i)}{1.05^{S(i)}}$)  & 52.866 & 0    & 1033\\
   $Q_2^2$ ($\frac{U(i)}{S(i)}$)          & 52.456 & 4    & 1025\\
   $Q_3^{1.025}$ ($\frac{U(i) * D(i)}{1.025^{S(i)}}$) & 49.437 & 0    & 966 \\
   $Q_1^2$ ($\frac{U(i)^2}{S(i)}$)         & 49.437 & 8    & 966 \\
   $Q_1^0$ ($\frac{D(i)^2}{S(i)}$)         & 46.469 & 3    & 908 \\
   $Q_2^0$ ($\frac{D(i)}{S(i)}$)          & 44.882 & 1    & 877 \\
\end{tabular}
\caption{\label{t:seqeval}Success rate of $Q_{i}$ depending on the quality formula.}
\end{table}
\renewcommand{\arraystretch}{1.2}

In case of \epcllemma three main strategies of creating a FOF trace from an inference trace were
considered. First, we tried to apply the \MESON translation of formulas. On one hand
this was most computationally expensive as it involves lambda-lifting and introducing the
apply functor, on the other hand it produces first-order formulas whose semantics are
closest to those of the higher-order formulas involved. Second, we tried to create arbitrary
FOF formulas of the same size as the one of the input HOL formula. Third, we modified the
second approach to also initialize \epcllemma with the already named theorems. The results
can be found in Table~\ref{t:epcleval}. The size of theorems is much more important than
the structure and initialization does not seem to help.

\begin{table}[htb]\centering
\begin{tabular}{cccc}\toprule
Added theorems & Success rate & Unique & Thms\\\midrule
 Preserve size & 55.732 & 15 & 1089 \\
 Preserve size and initialize   & 55.322 & 8  & 1081 \\
 \MESON translation & 47.339 & 11 & 925  \\\bottomrule
\end{tabular}
\caption{\label{t:epcleval}Success rate of \epcllemma depending on kinds of formulas given}
\end{table}

We next compare the versions of PageRank. The intersection between the first 10,000 lemmas
advised by $PR_1$ and $PR_2$ is 79\%, which suggests that the lemmas suggested by $PR_1$ are
already rather small. For the reverse PageRank it is the opposite: $PR_3$ and $PR_4$ have
only 11\% intersection. This makes the bigger lemmas suggested by $PR_3$ come out second
after the normalized combined $PR_6$ in Table~\ref{t:prankeval}.

\begin{table}[htb]\centering
\begin{tabular}{cccc}\toprule
Added theorems & Success rate & Unique & Thms\\\midrule
   $PR_6$ ($\frac{PR_1(i)+PR_3(i)}{S(i)}$) &53.173 &22  & 1039\\
   $PR_3$ (reverse $PR_1$)        &52.968 &13  & 1035\\
   $PR_5$ ($PR_1(i)+PR_3(i)$)	  &52.252 &14  & 1021\\
   $PR_2$ ($\frac{PR_1(i)}{S(i)}$)  &46.008 &5   & 899 \\
   $PR_4$ ($\frac{PR_3(i)}{S(i)}$)  &45.650 &8   & 892 \\
   $PR_1$        &42.272 &1   & 826 \\\bottomrule
\end{tabular}
\caption{\label{t:prankeval}Success rate of PageRank depending on  kinds of formulas given}
\end{table}

The resource-intensive \textit{fully-honest} evaluation is
limited to a relatively small subset of the core \HOLLight theorems,
however it confirms the \textit{almost-honest} results. While the
original success rate was 61.7\% (less than 14 methods are needed to
reach it), the success rate with lemma mining went up to 64.8\%
(again, less than 14 methods are needed). This means that the
non-cheating lemma-mining approaches so far improve the overall
performance of the AI/ATP methods over core \HOLLight by about 5\%.
The best method in the \textit{fully-honest} evaluation is $Q_2$ which
solves 46.2\% of the original problems when using 512 premises,
followed by $EQ_2$ (using the longest inference chain instead of $D$),
which solves 44.6 problems also with 512 premises. The best
PageRank-based method is $PR_2$ (PageRank divided by size), solving
41.4\% problems with 128 premises.

An interesting middle-way between the cheating and non-cheating
scenarios is the \textit{chained-conjecturing} evaluation, which
indicates the possible improvement when guessing good lemmas that are
``in the middle'' of long proofs. Since this is also quite expensive,
only the best lemma-mining method ($Q_2$) was evaluated on the
\HOLLight TRACE2. $Q_2$ itself solves (altogether, using different numbers of premises) 54.5\%
(1066) of the problems. This goes up to 61.4\% (1200 theorems) when
using only $chain_1$\textit{-conjecturing} and to 63.8\% (1247
theorems) when allowing also $chain_2$ and
$chain_3$\textit{-conjecturing}. These are 12.6\% and
17.0\% improvements respectively, see Table~\ref{t:chaineval}.

\begin{table}[htb]\centering
\begin{tabular}{cccc}\toprule
Length of chains & Success rate & Unique & Thms\\\midrule
-	&54.5	&519	&	1066	\\
1	&32.0	&75	&	627	\\
2	&12.2	&30	&	239	\\
3	&2.3	&12	&       46\\
4	&1.1	&4	& 	22\\
5	&0.3	&4	&	6\\
6	&0.3	&4	&	6\\
$>6$	&0.1	&2	& 	2\\
Total   &64.6   &       &       1264\\\bottomrule
\end{tabular}
\caption{\label{t:chaineval}Theorems found with chains of given lengths}
\end{table}

\subsection{Evaluation on \Flyspeck }

For the whole \Flyspeck the evaluation is due to the sizes of the data
limited to the tactical trace and the almost-honest scenario.
Table~\ref{t:flymethods} (the \Flyspeck counterpart of
Table~\ref{t:methods}) presents the performance of the various
lemma-based methods on the 1796 selected \Flyspeck theorems, together with the comparison and combination
with the old experiments. The combination of the 14 best methods tested here solves 37.6\% problems, 
and the combination of all methods solves 40.8\% problems. When combined with the 
most useful old methods developed in~\citep{holyhammer}, the performance of the
 best 14 methods is 44.2\%, i.e., we get a 21.4\% improvement over the older methods. 
The sequence of these 14 most-contributing methods is shown in Table~\ref{t:flygreedy}.

\begin{table}[htb]\centering
\begin{tabular}{cccc}\toprule
Strategy & Theorems (\%) & Unique & Theorems \\\midrule
 $PR_{1..5}$ (pagerank, sec.~\ref{sec:qpagerank})  & 36.860& 39	& 662\\
 $Q_{1..3}$ (direct quality, sec.~\ref{sec:qdirect}) & 35.913& 31	& 645\\
 $MC_{1..2}$ (graph cut, sec.~\ref{sec:qcut})  & 30.178& 1	& 542\\
 $EQ_{1..2}$ (\epcllemma, sec.~\ref{sec:qepcl}) & 29.677& 0	& 533\\
 all lemmas & 21.047& 26	& 378\\
 only named &  28.786 & 1 & 517\\
  14 best &37.584& & 675\\
   total   &40.813& & 733\\
\midrule
 \HH (14 best) & 36.414 & 127 & 654\\
  combined 14 best &44.209 & &	794\\
   total   &  47.884 & & 860\\
\bottomrule
 \end{tabular}
\caption{\label{t:flymethods}Comparison of the methods evaluated on the tactical trace
and the 1796 \Flyspeck theorems}
\end{table}

There are several issues related to the previous evaluations that need
explanation. First, the final 14-method \HH performance reported
in~\citep{holyhammer} was 39\%, while here it is only 36.4\%. The 39\%
were measured on the whole older version of \Flyspeck, while the
36.4\% here is the performance of the old methods limited to the 1796
problems selected from the set of 10,779 theorems that are preserved
between the old and the new version of \Flyspeck. Additionally, we
have recently reported~\citep{EasyChair:74} an improvement of the 39\%
performance to 47\% by using better learning methods and better E
strategies. However, that preliminary evaluation has been so far done
only on a smaller random subset of the old \Flyspeck, so we do not yet
have the corresponding data for all the 10,779 preserved theorems and
their 1796-big subselection used here for the comparison. A very rough
extrapolation is that the 47\% performance on the smaller subset will drop to
45\% on the whole old \Flyspeck, which when proportionally decreased by the
performance decrease of the old methods ($39/36.4$) yields 42\%
performance estimate on the new 1796-big set.  Third, we should note that the
new lemma-based methods are so far based only on learning from the ITP
(human-proof) dependencies, which is for \Flyspeck quite inferior to
learning on the dependencies extracted from minimized ATP proofs of
the problems. Fourth, we do use here the (one) best predictor and the ATP strategies developed in~\citep{EasyChair:74}, 
however, we have not so far explored and globally
optimized as many parameters (learners, features and their weightings,
premise slices, and ATP strategies) as we have done for the older
non-lemma methods; such global optimization is future work. 

\begin{table}[htb]\centering
\begin{tabular}{cccccccc}\toprule
Strategy & Pred. & Feat. & Lemmas & Prem. & ATP & Success & Thms\\\midrule
\HH	&NBayes	&typed, notriv& ATP-deps&154&epar &24.666 &443\\
$MC_2$	&k-NN	&typed & 1,000 lemmas 	&128&epar&	31.180&	560\\
All lemmas &k-NN&types & all lemmas 	&32&z3 &34.855 &626	\\
\HH	&NBayes	&types, notriv& ATP-deps&1024&epar &36.693 &659	\\
$Q_1$	&k-NN	&types & 60,000 lemmas 	&32&z3 &38.474 &691		\\
\HH	&NBayes	&typed, notriv& ATP-deps&92&vam &40.033 &719	\\
Only Named &k-NN&types & - 		&512&epar &40.980 &736		\\
$Q_1^2$	&k-NN	&types & 60,000 lemmas 	&32&z3 &41.759 &750		\\
\HH	&k-NN160&types, notriv&ATP deps &512&z3 &42.316 &760	\\
$Q^0_1$	&k-NN	&types & 60,000 lemmas 	&32&z3 &42.762 &768			\\
$PR_6$	&k-NN	&types & 20,000 lemmas 	&512&epar &43.207 &776			\\
\HH	&NBayes	&fixed & Human deps 	&512&epar &43.541 &782		\\
$PR_1$	&k-NN	&types & 20,000 lemmas 	&32&z3 &43.875 &788				\\
$PR_4$	&k-NN	&types & 20,000 lemmas 	&128&epar &44.209 &794			\\
\bottomrule
\end{tabular}
\caption{\label{t:flygreedy}Combined 14 best covering sequence}
\end{table}

So while the 21.4\% improvement over~\citep{holyhammer} is valid, a
full-scale evaluation of all the methods on the whole new \Flyspeck\footnote{Such evaluation could take another month with our current resources.}
will likely show a smaller improvement due to the lemma-mining
methods. A very conservative estimate is again 5\% (44.2\%/42\%),
however a much more realistic is probably 10\%, because the effect of
learning from ATP proofs is quite significant. Higher lemma-based performance on \Flyspeck 
than on the core \HOLLight is quite plausible: the core \HOLLight library is much smaller, more stable and optimized, while \Flyspeck is a fast-moving project written by several authors, and the library structuring there is more challenging.

As expected the graph cutting method ($MC$) does indeed produce the smallest dependency graph passed to
the predictors. For 10,000
added lemmas the average number of edges in the $MC$-produced dependency graph is 37.0, compared
with the average over all strategies being 42.9 dependencies per theorem and \epcllemma producing
graphs with the biggest number: 63.1 dependencies. This however does not yet correspond to high
success rates in the evaluation, possibly due to the fact that graph cutting does not so far 
take into account the number of small steps needed to prove the added lemma. On the other hand, Table~\ref{t:flygreedy}
shows that graph cutting provides the most complementary method, adding about 25\% more new solutions to the best method available.

\begin{table}[htb]\centering
\begin{tabular}{cccc}\toprule
Added lemmas & Theorems (\%) & Unique & Theorems \\\midrule
60,000	&36.804	&37	&661\\
20,000	&35.523	&18	&638\\
10,000	&33.463	&3	&601\\
0	&28.786	&1	&517\\
5,000	&27.951	&0	&502\\
1,000	&27.895	&0	&501\\
all	&21.047	&26	&378\\
\bottomrule
\end{tabular}
\caption{\label{t:addedlemmas}}
\end{table}
Finally we analyze the influence of the number of added lemmas on the success rate in Table~\ref{t:addedlemmas}.
As expected adding more lemmas does improve the general performance up to a certain point. The experiments performed
with all the lemmas added are already the weakest. However, when it comes to the problems solved only with a certain
number of lemmas added, using all the lemmas comes out complementary to the other numbers.

\subsection{Examples}

We have briefly looked at some first examples of the problems that can
be solved only with the lemma-based methods. So far we have detected
two main effects how such new proofs are achieved: (i) the new lemma
(or lemmas) is an easy-but-important specialization of a general
theorem or theory, either directing the proof search better than its
parents or just working better with the other premises, and (ii) no new lemma is needed, but learning on the newly
added lemmas improves the predicting systems, which then produce
better advice for a previously unsolvable problem. The second effect
is however hard to exemplify, since the number of alternative
predictions we tried is high, and it usually is not clear why a particular prediction did not succeed. 
An example in the first category is the theorem

\begin{holnb}\holthm{

AFFINE_ALT: |- affine s <=> (!x y u. x IN s /\ y IN s ==> (&1 - u) 

}\end{holnb}
which E can prove using 15 premises, three of them being new lemmas that are quite ``trivial'' consequences of more general theorems:
\begin{holnb}\holthm{

NEWDEP309638: |- &1 - a + a = &1
NEWDEP310357: |- -- &1 * -- &1 = &1
NEWDEP272099_conjunct1: |- !m. &m + -- &m = &0

}\end{holnb}
Another example in the first category is theorem
\begin{holnb}\holthm{

MEASURABLE_ON_NEG: |- !f s. measurable_on f s ==> measurable_on (\textbackslash{x}. --f x) s

}\end{holnb}
whose proof uses a few basic vector facts plus one added lemma:
\begin{holnb}\holthm{

NEWDEP1643063: measurable_on f s |- measurable_on ((

}\end{holnb}
This lemma appeared in the proof of the close theorem
\begin{holnb}\holthm{

MEASURABLE_ON_CMUL: |- !c f s. measurable_on f s ==> measurable_on (\textbackslash{x}. c 

}\end{holnb}
The lemma here is almost the same as the theorem where it was first used, but it likely works better in the FOF encoding because the
lambda function is eliminated.

\section{Future Work and Conclusion}
\label{Future}

We have proposed, implemented and evaluated several approaches that try to efficiently
find the best lemmas and re-organize a large corpus of computer-understandable human
mathematical ideas, using the millions of logical dependencies between
the corpus' atomic elements. We believe that such conceptual
re-organization is a very interesting AI topic that is best studied in the context of large, fully
semantic corpora such as \HOLLight and \Flyspeck. 
The byproduct of this work
are the exporting and post-processing techniques resulting in the
publicly available proof graphs that can serve as a basis for further
research.

The most conservative improvement in the strength of automated
reasoning obtained so far over the core \HOLLight thanks to lemma
mining is about 5\%. The improvement in the strength of automated
reasoning obtained over \Flyspeck problems is 21.4\% in comparison to
the methods developed in~\citep{holyhammer}, however this improvement
is not only due to the lemma-mining methods, but also due to some of
the learning and strategy improvements introduced
in~\citep{EasyChair:74}. A further large-scale evaluation using
learning from ATP proofs and global parameter optimization is needed
to exactly measure the contribution and overall strength of the
various AI/ATP methods over the whole \Flyspeck corpus.

There are many further directions for this work. The lemma-mining methods
can be made faster and more incremental, so that the lemma quality is not
completely recomputed after a lemma is named. Fast PageRank-based clustering
should be efficiently implemented and possibly combined with the other methods
used. ATP-style normalizations such as subsumption need to be correctly merged
with the detailed level of inferences used by the \HOLLight proof graph. 
Guessing of good intermediate lemmas for proving harder theorems is an obvious
next step, the value of which has already been established to a certain extent in
this work.

\section{Acknowledgments}

We would like to thank Stephan Schulz for help with running \epcllemma,
Yury Puzis and Geoff Sutcliffe for their help with the \Agint tool and
Ji\v{r}\'i Vysko\v{c}il and Petr Pudl\'ak for many discussions about
extracting interesting lemmas from proofs.

\bibliography{ate11}
\bibliographystyle{elsart-harv}

\end{document}